\documentclass[oneside,11pt]{article}

\usepackage{color}
\usepackage{graphicx}
\usepackage{amsmath}
\usepackage{algorithm}
\usepackage[noend]{algpseudocode}
\usepackage{float}
\usepackage[utf8]{inputenc}
\usepackage[titletoc,title]{appendix}

\usepackage[a4paper, total={6in, 9in}]{geometry}
\usepackage[english]{babel}
\usepackage[backend=biber]{biblatex}

\usepackage{dirtytalk}

\title{Classifying movie genres by analyzing text reviews}
\author{Adam Nyberg}
\date{ }

\begin{document}

\maketitle

\begin{abstract}
This paper proposes a method for classifying movie genres by only looking at text reviews. The data used are from \textit{Large Movie Review Dataset v1.0} \cite{maas-EtAl:2011:ACL-HLT2011} and IMDb. This paper compared a K-nearest neighbors (KNN) model and a multilayer perceptron (MLP) that uses tf-idf as input features. The paper also discusses different evaluation metrics used when doing multi-label classification. For the data used in this research, the KNN model performed the best with an $accuracy$ of $55.4\%$ and a $Hamming\ loss$ of $0.047$.
\end{abstract}

\section{Introduction}
By only reading a single text review of a movie it can be difficult to say what the genre of that movie is, but by using text mining techniques on thousands of movie reviews is it possible to predict the genre? 

This paper explores the possibility of classifying genres of a movie based only on a text review of that movie. This is an interesting problem because to the naked eye it may seem difficult to predict the genre by only looking at a text review. One example of a review can be seen in the following example:

\say{I liked the film. Some of the action scenes were very interesting, tense and well done. I especially liked the opening scene which had a semi truck in it. A very tense action scene that seemed well done. Some of the transitional scenes were filmed in interesting ways such as time lapse photography, unusual colors, or interesting angles. Also the film is funny is several parts. I also liked how the evil guy was portrayed too. I'd give the film an 8 out of 10.}\footnotemark

\footnotetext{http://www.imdb.com/title/tt0211938/reviews}

From the quoted review, one could probably predict the movie falls in the action genre; however, it would be difficult to predict all three of the genres (action, comedy, crime) that International Movie Database (IMDB) lists. With the use of text mining techniques it is  feasible to predict multiple genres based on a review.

There are numerous previous works on classifying the sentiment of reviews, e.g., \citetitle{maas-EtAl:2011:ACL-HLT2011} by \citeauthor{maas-EtAl:2011:ACL-HLT2011}. There are fewer scientific papers available on specifically classifying movie genres based on reviews; therefore, inspiration for this paper comes from papers describing classification of text for other or general contexts. One of those papers is \citetitle{DBLP:journals/corr/cmp-lg-9707002} where \citeauthor{DBLP:journals/corr/cmp-lg-9707002} describe how to use a multilayer perceptron (MLP) for genre classification.

All data, in the form of reviews and genres, used in this paper originates from IMDb.

\section{Theory}
In this section all relevant theory and methodology is described. Table \ref{table:terms} lists basic terminology and a short description of their meaning.

\begin{table}[ht!]
\centering
\begin{tabular}{|c c|} 
 \hline
 Name & Description \\
 \hline\hline
 Corpus & A structured set of documents.\\
 Document & A list of tokens. In this paper a document refers to one review. \\
 Term & A word. \\
 \hline
\end{tabular}
\caption{List of basic terminology.}
\label{table:terms}
\end{table}

\subsection{Preprocessing}
Data preprocessing is important when working with text data because it can reduce the number of features and it formats the data into the desired form \cite{hotho2005brief}.

\subsubsection{Stop words}
Removing stop words is a common type of filtering in text mining. Stop words are words that usually contain little or no information by itself and therefore it is better to remove them. Generally words that occur often can be considered stop words such as \textit{the}, \textit{a} and \textit{it}. \cite{hotho2005brief}

\subsubsection{Lemmatization}
Lemmatization is the process of converting verbs into their infinitive tense form and nouns into their singular form. The reason for doing this is to reduce words into their basic forms and thus simplify the data. For example \textit{am}, \textit{are} and \textit{is} are converted to \textit{be}. \cite{hotho2005brief}

\subsubsection{Tf-idf}
A way of representing a large corpus is to calculate the Term Frequency Inverse Document Frequency (tf-idf) of the corpus and then feed the models the tf-idf. As described in \citetitle{ramos2003using} by \citeauthor{ramos2003using} tf-idf is both efficient and simple for matching a query of words with a document in a corpus. Tf-idf is calculated by multiplying the Term Frequency (tf) with the Inverse Document Frequency (idf) , which is formulated as

\begin{equation}
tfidf(t,d,D) = tf(t,d) \cdot idf(t,D)
\end{equation}

where $d$ is a document in corpus $D$ and $t$ is a term. $tf(t,d)$ is defined as

\begin{equation}
td(t,d) = 0.5 + 0.5 \cdot \frac{f_{t,d}}{max\{f_{t^{'},d}:t^{'} \in d \}}
\end{equation}

and $idf(t,D)$ is defined as

\begin{equation}
idf(t,D) = log \frac{N}{|\{d \in D : t \in d\}|}
\end{equation} 

where $f_{t,d}$ is the number of times $t$ occurs in $d$ and $N$ total number of documents in the corpus.

\subsection{Models}
\subsubsection{Multilayer Perceptron}
MLP is a class of feedforward neural network built up by a layered acyclic graph.
An MLP consists of at least three layers and non-linear activations. The first layer is called input layer, the second layer is called hidden layer and the third layer is called output layer. The three layers are fully connected which means that every node in the hidden layer is connected to every node in the other layers. MLP is trained using backpropagation, where the weights are updated by calculating the gradient descent with respect to an error function. \cite{kotsiantis2007supervised}

\subsubsection{K-nearest Neighbors}
K-nearest Neighbors (KNN) works by evaluating similarities between entities, where $k$ stands for how many neighbors are taken into account during the classification. KNN is different from MLP in the sense that it does not require a computationally heavy training step; instead, all of the computation is done at the classification step. There are multiple ways of calculating the similarity, one way is to calculate the Minkowski distance. The Minkowski distance between two points 

\begin{equation}
X=(x_1,x_2,...,x_n)
\end{equation} 

and 

\begin{equation}
Y=(y_1,y_2,...,y_n)
\end{equation}

is defined by

\begin{equation}
D(X,Y)=(\sum\limits_{i=1}^n|x_i-y_i|^p)^\frac{1}{p}
\end{equation} 

where $p=2$ which is equal to the Euclidean distance. \cite{hotho2005brief}

\subsection{Evaluation}
\label{theory:eval}
When evaluating classifiers it is common to use accuracy, precision and recall as well as Hamming loss. Accuracy, precision and recall are defined by the the four terms true positive ($TP$), true negative ($TN$), false positive ($FP$) and false negative ($FN$) which can be seen in table \ref{table:confusion}.

\begin{table}[ht!]
\centering
\begin{tabular}{ r|c|c| }
\multicolumn{1}{r}{}
 &  \multicolumn{1}{c}{Actual true}
 & \multicolumn{1}{c}{Actual false} \\
\cline{2-3}
Predicted true & $TP$ & $FP$ \\
\cline{2-3}
Predicted false & $FN$ & $TN$ \\
\cline{2-3}

\end{tabular}
\caption{Definition of $TP$, $TN$, $FP$ and $FN$.}
\label{table:confusion}
\end{table}

Accuracy is a measurement of how correct a model's predictions are and is defined as 

\begin{equation}
accuracy =  \frac{TP + TN}{TP + TN + FP + FN}
\label{eq:acc}
\end{equation}.

Precision is a ratio of how often positive predictions actually are positve and is defined as

\begin{equation}
precision = \frac{TP}{TP + FP}
\label{eq:precision}
\end{equation}.

Recall is a measurement of how good the model is to find all true positives and is defined as

\begin{equation}
Recall =  \frac{TP}{TP + FN}
\label{eq:recall}
\end{equation}. \cite{powers2011evaluation}

It has been shown that when calculating precision and recall on multi-label classifiers, it can be advantageous to use micro averaged precision and recall \cite{6471714}. The formulas for micro averaged precision are expressed as

\begin{equation}
precision_{micro} = \frac{\sum_{i=1}^{K} TP_i}{\sum_{i=1}^{K} TP_i \sum_{i=1}^{K} FP_i}
\label{eq:precision_micro}
\end{equation}

\begin{equation}
recall_{micro} = \frac{\sum_{i=1}^{K} TP_i}{\sum_{i=1}^{K} TP_i \sum_{i=1}^{K} FN_i}
\label{eq:recall_micro}
\end{equation}

where $i$ is label index and $K$ is number of labels.

Hamming loss is different in the sense that it is a loss and it is defined as the fraction of wrong labels to the total number of labels. Hamming loss can be a good measurement when it comes to evaluating multi-label classifiers. the hamming loss is expressed as

\begin{equation}
Hamming\ loss = \frac{1}{N \cdot K} \sum\limits_{i=1}^N\sum\limits_{j=1}^K xor(y_{i,j},z_{i,j})
\end{equation} 

where $N$ is number of documents, $K$ number of labels, $y_{i,j}$ is the target value and $z_{i,j}$ is predicted value. \cite{sorower2010literature}

\section{Data}
Data used in this paper comes from two separate sources. The first source was \textit{Large Movie Review Dataset v1.0} \cite{maas-EtAl:2011:ACL-HLT2011} which is a dataset for binary sentiment analysis of moview reviews. The dataset contains a total of $50 000$ reviews in raw text together with information on whether the review is positive or negative and a URL to the movie on IMDb. The sentiment information was not used in this paper. Out of the $50 000$, reviews only $7 000$ were used because of limitations on computational power, resulting in a corpus of $7 000$ documents.

The second source of data was the genres for all reviews which were scraped from the IMDb site. A total of 27 different genres were scraped. A list of all genres can be find in Appendix \ref{appendix:genres}. A review can have one genre or multiple genres. For example a review can be for a movie that is both \textit{Action}, \textit{Drama} and \textit{Thriller} at the same time while another move only falls into \textit{Drama}. 

\section{Method}
This section presents all steps needed to reproduce the results presented in this paper. 

\subsection{Data collection}
In this paper the data comes from two sources where the first is a collection of text reviews. Those reviews were downloaded from \textit{Large Movie Review Datasets} website \footnotemark. Because only $7 000$ reviews was used in this paper all of them were from the `train` folder and split evenly between positive reviews and negative reviews.

The genres for the reviews where obtained by iterating through all reviews and doing the following steps:

\begin{enumerate}
  \item Save the text of the review.
  \item Retrieve IMDb URL to the movie from the \textit{Large Movie Review Datasets} data.
  \item Scrape that movie website for all genres and download the genres.
\end{enumerate}

\begin{figure}[!ht]
  \centering
  \includegraphics[width=0.8\textwidth]{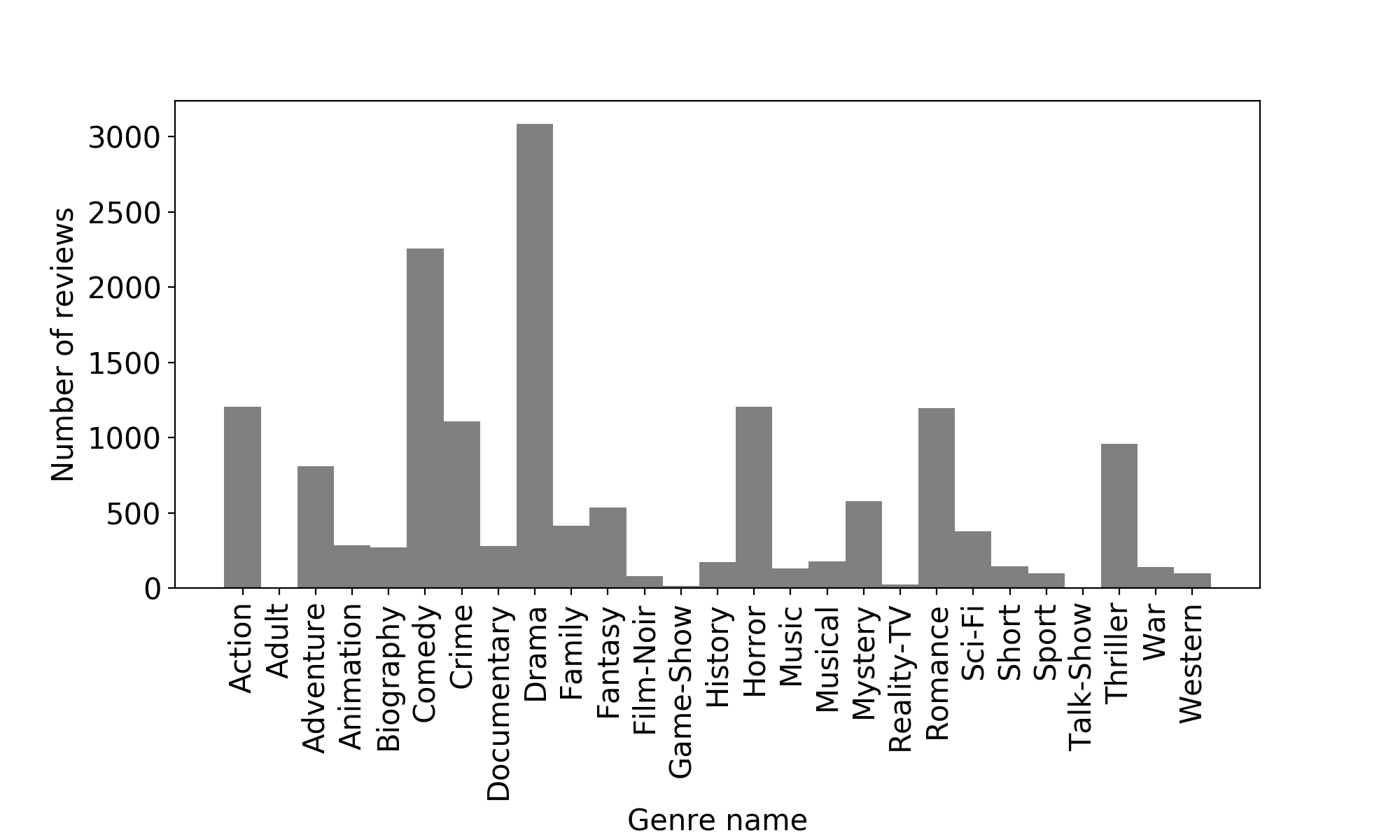}
  \caption{Histogram showing the distribution of genres.}
  \label{fig:totaldist}
\end{figure}

The distribution of genres was plotted in a histogram to check that the scraped data looked reasonable and can be seen in figure \ref{fig:totaldist}. All genres with less than $50$ reviews corresponding to that genre were removed.

\begin{figure}[!ht]
  \centering
  \includegraphics[width=0.8\textwidth]{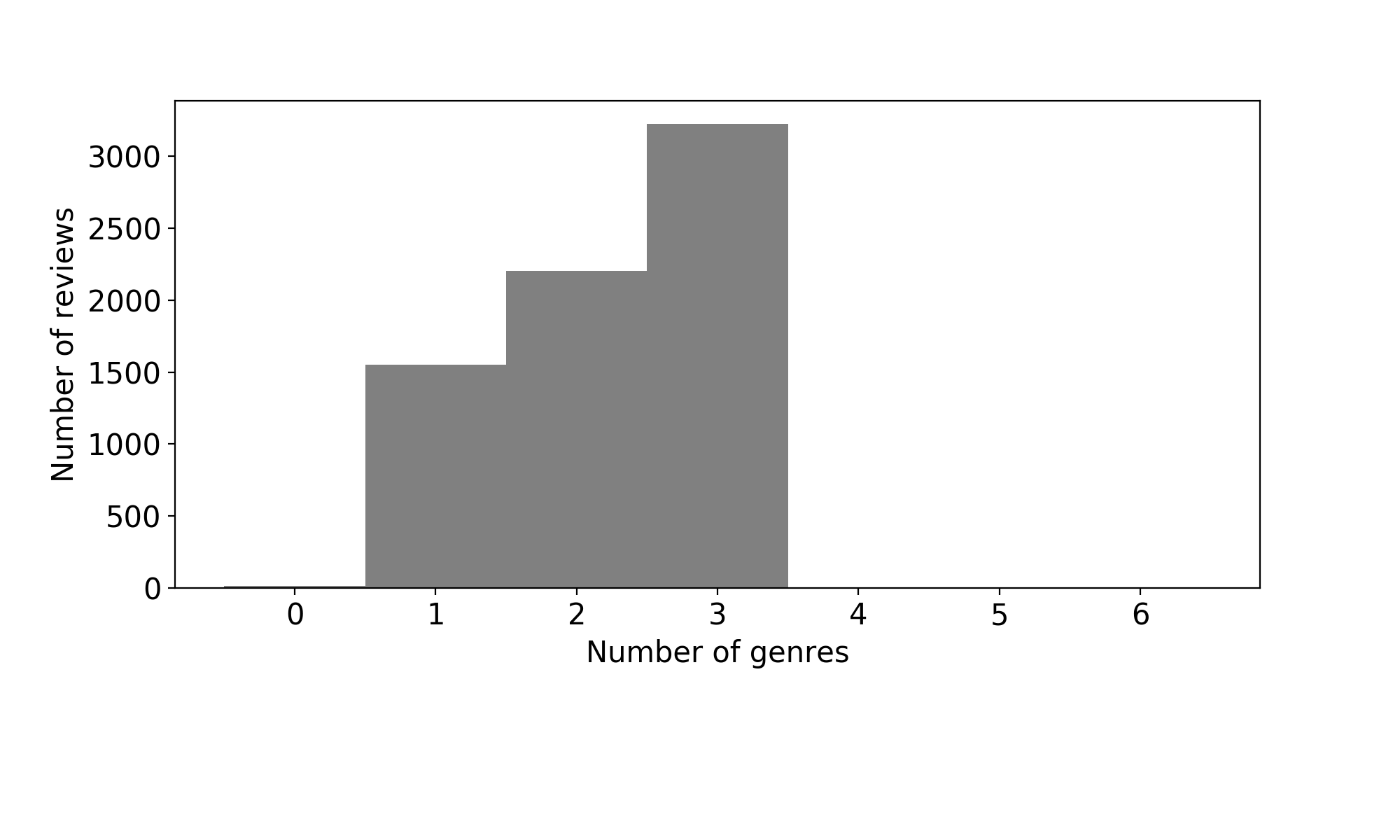}
  \caption{Histogram showing the distribution of genres per review.}
  \label{fig:genres-per-review}
\end{figure}

The number of genres per review can be seen in figure \ref{fig:genres-per-review} and it shows that it is most common for a review to have three different genres; furthermore, it shows that no review has more than three genres.

\footnotetext{http://ai.stanford.edu/~amaas/data/sentiment}

\subsection{Data preprocessing}
All reviews were preprocessed according to the following steps:

\begin{enumerate}
  \item Remove all non-alphanumeric characters.
  \item Lower case all tokens.
  \item Remove all stopwords.
  \item Lemmatize all tokens.
\end{enumerate}

Both the removal of stopwords and lemmatization were done with Python's\footnotemark\ Natural Language Toolkit (NLTK)\footnotemark. Next the reviews and corresponding genres were split into a training set and a test set with $70\%$ devided into the train set and $30\%$ into the test set.

The preprocessed corpus was then used to calculate a tf-idf representing all reviews. The calculation of the tf-idf was done using scikit-learn's\footnotemark module \textit{TfidfVectorizer}. Both transform and fit were run on the training set and only the transform was run on the test set. The decision to use tf-idf as a data representation is supported by \citeauthor{ramos2003using} in \citetitle{ramos2003using} which concludes that tf-idf is both simple and effective at categorizing relevant words.

\footnotetext{https://www.python.org}
\footnotetext{http://www.nltk.org}
\footnotetext{http://scikit-learn.org}

\subsection{Model}
This paper experimented with two different models and compared them against each other. The inspiration for the first model comes from \citeauthor{DBLP:journals/corr/cmp-lg-9707002} in their paper \citetitle{DBLP:journals/corr/cmp-lg-9707002} where they used an MLP for text genre detection. The model used in this paper comes from scikit-learn's \textit{neural\_network} module and is called \textit{MLPClassifier}. Table \ref{table:MLPparams} shows all parameters that were changed from the default values.

\begin{table}[ht!]
\centering
\begin{tabular}{ |c | c| }
\hline
\textbf{Parameter} & \textbf{Value}  \\
solver & lbfgs \\
\hline
\end{tabular}
\caption{Values of non-default parameters for the MLP model.}
\label{table:MLPparams}
\end{table}

The second model was a KNN which was chosen because of it is simple and does not require the pre-training that the MLP needs. The implementation of this model comes from scikit-learn's \textit{neighbors} module and is called \textit{KNeighborsClassifier}. The only parameter that was changed after some trial and error was the k-parameter which was set to $3$.

Both models were fitted using the train set and then predictions were done for the test set.

\subsection{Evaluation}
For evaluation the $accuracy$ and $Hamming\ loss$ was calculated as defined in section \ref{theory:eval} for both the MLP model and the KNN model. For precision and recall formulas \ref{eq:precision_micro} and \ref{eq:recall_micro} were used because of their advantage in multi-label classification. The distribution of predicted genres was also shown in a histogram and compared to the target distribution of genres. 

Furthermore the ratio of reviews that got zero genres predicted was also calculated and can be expressed as

\begin{equation}
No\ genre\ ratio = \frac{M}{z_{total}}
\label{eq:recall_micro}
\end{equation}

where $M$ is the number of reviews without any predicted genre and $z_{total}$ is the total amount of predicted reviews.

\section{Result}
Table \ref{table:results} shows the $accuracy$, $precision_{micro}$ and $recall_{micro}$ for the models. The KNN model had a higher accuracy of $0.554$ compared to MPL's accuracy of $0.37$ and the KNN model had a higher recall but slightly lower precision than the MLP model.

\begin{table}[ht!]
\centering
\begin{tabular}{ |c|c|c|c| }
\hline
\textbf{Model} & \textbf{Accuracy} & \textbf{Precision micro} & \textbf{Recall micro}  \\
\hline
\hline
MLP & $0.37$ & $0.816$ & $0.597$ \\
\hline
KNN & $0.554$ & $0.807$ & $0.677$ \\
\hline
\end{tabular}
\caption{$accuracy$, $precision_{micro}$ and $recall_{micro}$ for the models.}
\label{table:results}
\end{table}

Table \ref{table:results2} shows the $Hamming\ loss$ and $No\ genre\ ratio$ for the models, it shows that the KNN model had lower values for both the $Hamming\ loss$ and $No\ genre\ ratio$ compared to the MLP model.

\begin{table}[ht!]
\centering
\begin{tabular}{ |c|c|c| }
\hline
\textbf{Model} & \textbf{Hamming loss} & \textbf{No genre ratio} \\
\hline
\hline
MLP & $0.051$ & $0.086$ \\
\hline
KNN & $0.047$ & $0.053$ \\
\hline
\end{tabular}
\caption{$Hamming\ loss$ and $No\ genre\ ratio$ for the models.}
\label{table:results2}
\end{table}

\begin{figure}[!ht]
  \centering
  \includegraphics[width=0.8\textwidth]{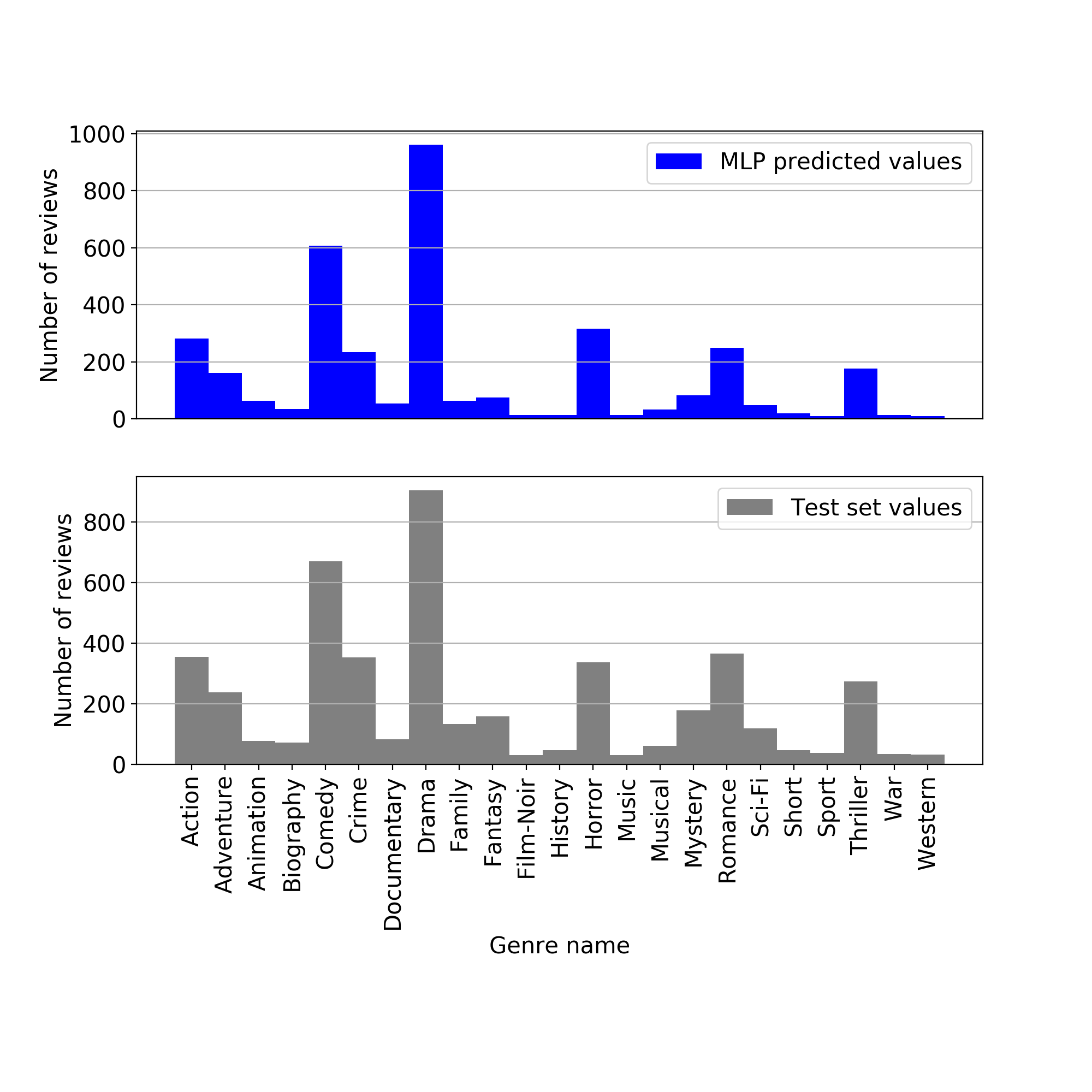}
  \caption{Distribution of genres in MLP predictions and test set.}
  \label{fig:dist-mlp-comp}
\end{figure}

\begin{figure}[!ht]
  \centering
  \includegraphics[width=0.8\textwidth]{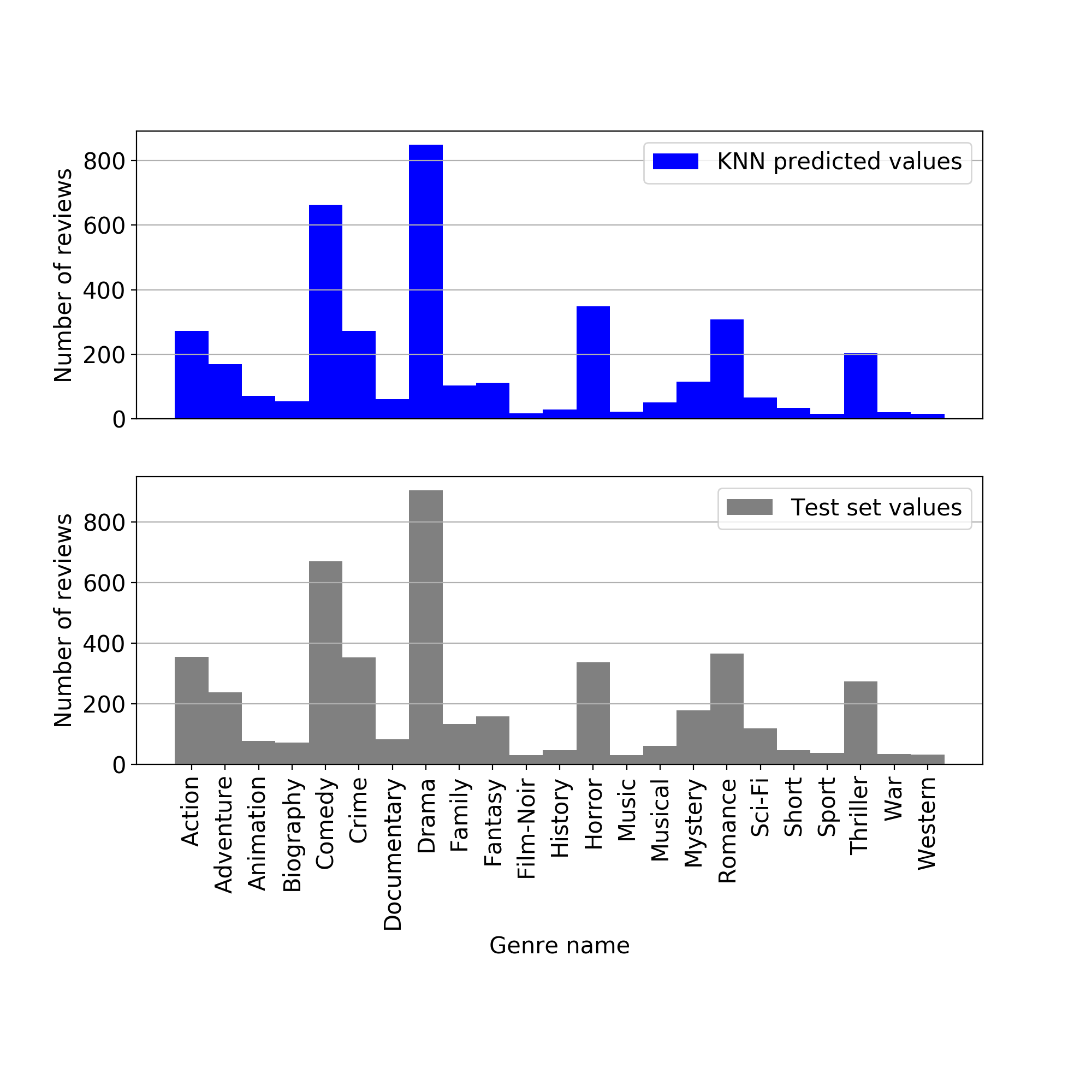}
  \caption{Distribution of genres in KNN predictions and test set.}
  \label{fig:dist-knn-comp}
\end{figure}

Figure \ref{fig:dist-mlp-comp} shows the distribution of the genres for the predicted values when using MLP and the test set. The same comparison between KNN and the test set can be seen in figure \ref{fig:dist-knn-comp}.

\section{Discussion}
When looking at the results it is apparent that KNN is better than MLP in these experiments. In particular, the $accuracy$ stands out between KNN and MLP where KNN got $0.554$ and MLP got $0.37$ which is considered a significant difference. Given that the $precision_{micro}$ was relatively high for both models, this result hints that the models only predicted genres when the confidence was high, which resulted in fewer genres being predicted than the target. This can also be confirmed by looking at the figures \ref{fig:dist-mlp-comp} and \ref{fig:dist-knn-comp} where the absolute number of reviews predicted for most genres was lower than the target. This unsatisfyingly low $accuracy$ can be explained by the multi-label nature of the problem in this paper. Even if the model correctly predicted 2 out of three genres it is considered a misclassification. A reason for the low accuracy could be that the models appeared to be on the conservative side when predicting genres.

Another factor that affected the performance of the models was the $No\ genre\ ratio$ which confirmed that over $5\%$ of the reviews for the KNN model and over $8\%$ of the reviews for the MLP model did not receive any predicted genre. Because no review had zero genres all predictions with zero genres are misclassified and this could be a good place to start when improving the models.

Furthermore, when looking at the $Hamming\ loss$ it shows that when looking at the individual genres for all reviews the number of wrong predictions are very low which is promising when trying to answer this paper's main question: whether it is possible to predict the genre of the movie associated with a text review. It should be taken into account that this paper only investigated about $7 000$ movie reviews and the results could change significantly, for better or for worse, if a much larger data set was used. In this paper, some of the genres had very low amounts of training data, which could be why those genres were not predicted in the same frequency as the target. An example of that can be seen by looking at genre \textit{Sci-Fi} in figure \ref{fig:dist-mlp-comp}.

\section{Conclusion}
This paper demonstrates that by only looking at text reviews of a movie, there is enough information to predict its genre with an $accuracy$ of $0.554$. This result implies that movie reviews carry latent information about genres. This paper also shows the complexity of doing prediction on multi-label problems, both in implementation and data processing but also when it comes to evaluation. Regular metrics typically work, but they mask the entire picture and the depth of how good a model is.

Finally this paper provides an explanation of the whole process needed to conduct an experiment like this. The process includes downloading a data set, web scraping for extra information, data preprocessing, model tuning and evaluation of the results. 

\pagebreak

\printbibliography

\pagebreak

\begin{appendices}

\section{All genres}
\label{appendix:genres}
\texttt{Action\\ 
Adult\\ 
Adventure\\ 
Animation\\ 
Biography\\ 
Comedy\\ 
Crime\\ 
Documentary\\ 
Drama\\ 
Family\\ 
Fantasy\\ 
Film-Noir\\ 
Game-Show\\ 
History\\ 
Horror\\ 
Music\\ 
Musical\\ 
Mystery\\ 
Reality-TV\\ 
Romance\\ 
Sci-Fi\\ 
Short\\ 
Sport\\ 
Talk-Show\\ 
Thriller\\ 
War\\ 
Western}

\end{appendices}

\end{document}